\documentclass{article}
\usepackage{arxiv}
\usepackage[utf8]{inputenc} 
\usepackage[T1]{fontenc}    
\usepackage{url}            
\usepackage{booktabs}       
\usepackage{amsfonts}       
\usepackage{nicefrac}       
\usepackage{microtype}      
\usepackage{graphicx}
\usepackage[square,numbers]{natbib}
\usepackage{amsmath,amssymb}
\usepackage{listings}
\renewcommand{\headeright}{}
\renewcommand{\undertitle}{}
\newcommand{\Kp}{$K^{(p)}$}

\title{Implementing Rational Choice Functions with LLMs and Measuring their Alignment with User Preferences}

\author{Anna Karnysheva \\
Spoken Language Systems~(LSV) \\
Saarland University \\
Saarbrücken,
Germany \\
\texttt{akarnysheva@lsv.uni-saarland.de} \\
\And
Christian Drescher \\
Mercedes-Benz AG \\
Stuttgart,
Germany \\
\texttt{christian.d.drescher@mercedes-benz.com} \\
\And
Dietrich Klakow \\
Spoken Language Systems~(LSV) \\
Saarland University \\
Saarbrücken,
Germany \\
\texttt{dietrich.klakow@lsv.uni-saarland.de}
}

\begin{document}
\maketitle

\begin{abstract}
As large language models (LLMs) become integral to intelligent user interfaces (IUIs), their role as decision-making agents raises critical concerns about alignment. Although extensive research has addressed issues such as factuality, bias, and toxicity, comparatively little attention has been paid to measuring alignment to preferences, i.e., the relative desirability of different alternatives, a concept used in decision making, economics, and social choice theory. However, a reliable decision-making agent makes choices that align well with user preferences.

In this paper, we generalize existing methods that exploit LLMs for ranking alternative outcomes by addressing alignment with the broader and more flexible concept of user preferences, which includes both strict preferences and indifference among alternatives. To this end, we put forward design principles for using LLMs to implement rational choice functions, and provide the necessary tools to measure preference satisfaction. We demonstrate the applicability of our approach through an empirical study in a practical application of an IUI in the automotive domain. 
\end{abstract}

\section{Introduction}
In recent years, large language models~(LLMs) have revolutionized natural language processing~(NLP), advancing the state-of-the-art across a wide array of tasks. Beyond traditional NLP applications such as question answering and text classification, these models have demonstrated remarkable abilities in challenging domains such as reasoning~\cite{liu2023evaluating, openai2023gpt4, geminiteam2023gemini}, often exhibiting human-like proficiency~\cite{brown2020language}. As a result, LLMs are increasingly being integrated into intelligent user interfaces~(IUIs) to enhance user experiences by providing more intelligent, context-aware, and personalized interactions. However, as LLMs assume more critical roles in automating decisions and behaviors as intelligent agents, ensuring that their actions align with user preferences, values, and objectives becomes paramount. Misaligned artificial intelligence~(AI) agents may make decisions that are unreliable or untrustworthy, potentially compromising user trust and the effectiveness their applications.

Previous research on LLM alignment has focused primarily on specific dimensions such as factuality~\cite{honovich2022true}, bias~\cite{Santurkar23_bias, zhuo2023red}, and toxicity~\cite{zhuo2023red}. Although these efforts have advanced our understanding of LLM alignment, they do not address \emph{preferences}, i.e., the relative desirability of different outcomes, a concept widely utilized in decision-making, economics, and various fields of social choice theory.
However, selecting the most desirable from a set of alternatives in a manner that reflects user preferences is a fundamental aspect of intelligent decision making in IUIs. We can illustrate this with a simple example.

\textit{Example.}
Consider the task of proposing the most helpful item from a subset of objects consisting of an umbrella, a raincoat, a jacket, a laptop, and keys to protect themselves from precipitation (for example, rain). A human user would arguably prefer either an umbrella or a raincoat over a jacket, and all of these over a laptop or keys, while remaining indifferent between the umbrella and raincoat, and between laptop and keys, respectively. Suppose we have two intelligent agents to support this task: Agent~1 encodes the binary classification $\{\text{raincoat}, \text{umbrella}\} \succ \{\text{jacket}, \text{laptop}, \text{keys}\}$, while Agent~2 encodes the ranking $\text{raincoat} \succ \text{umbrella} \succ \text{jacket} \succ \text{laptop} \succ \text{keys}$. If only an umbrella and a jacket are available, both agents choose the item (umbrella) which aligns with the alternative preferred by the user. If, however, only a jacket and keys are available, however, Agent~1 is indifferent between the alternatives even though the user strictly prefers the jacket over keys. On the other hand, if both a raincoat and an umbrella are available, Agent~2 will always choose the raincoat, thus, for instance, reminding the user to consider a raincoat over an umbrella, despite the user being indifferent between them.

For an intelligent agent to work reliably across all subsets of alternatives, it must implement a choice function that aligns well with user preferences.
To examine the alignment of intelligent agents that make decisions based on the output of LLMs, we need to address two key challenges: (1)~extracting the preferences encoded by LLMs, and (2)~measuring alignment between those preferences and user preferences.

First, since LLMs do not encode preferences explicitly, we must infer them from the model's outputs. One straightforward approach is to query the model. For instance, related work in the field of information retrieval has demonstrated that a ranking among alternatives can be inferred from an LLM, e.g., by asking the model to provide a score to each alternative~\cite{ji2023}, where higher scores indicate higher relevance. Or, by presenting the model with pairs of alternatives and asking it to indicate the more relevant one~\cite{qin2023}. Alternatively, the model's generative capabilities can be leveraged to sequence the complete list of alternatives according to their relevance~ \cite{sun2023chatgpt, ma2023, hou2023}. While these existing methods provide rankings, they do not account for indifference between alternatives. Therefore, we need more general methods that leverage LLMs to implement choice functions capable of capturing both indifference and strict preferences, and consequently, more general evaluation metrics for assessing their alignment with user preferences. 

We aim to enhance the trustworthiness and effectiveness of LLMs within IUIs by ensuring they operate in accordance with user preferences. In this paper, we address \emph{preference orderings} - a very general class of preferences that can encode strict preferences like rankings, but, more importantly, also capturing indifference, expressing that alternatives are equally preferable. To this end, we propose a comprehensive framework. Our contributions are threefold:
\begin{enumerate}
\item \textbf{Design principles:} We introduce design principles to implement rational choice functions using LLMs, including existing ideas borrowed from related work~\cite{qin2023} that we generalize from ranking to capture preferences. \texttt{Pairwise-Score} uses pairwise comparisons to assign ranks with ties among alternatives, and \texttt{Pairwise-SCC} uses strongly connected components in a directed graph derived from pairwise prompting to capture preference relationships. \texttt{Pairwise-Test} implements the intuitive but na\"ive idea of simply querying an LLM without enforcing that choices are rational. Compatible related work includes \texttt{Pointwise}~\cite{ji2023} and \texttt{Listwise}~\cite{sun2023chatgpt,ma2023} methods. While \texttt{Pointwise} assigns a score to each alternative, \texttt{Listwise} considers all alternatives at once by producing a ranking. 

\item \textbf{Evaluation metrics for alignment assessment:} We put forward metrics tailored to measuring alignment with user preferences. \emph{Strict preference overlap} quantifies \emph{partial alignment}, focusing on strict preference adherence, while \emph{Kendall distance with penalty}, \Kp, assesses \emph{full alignment} across all preference relations, i.e., also including indifference between alternatives.

\item \textbf{Empirical validation:} We validate the proposed framework through an empirical study in the automotive domain, where an IUI makes decisions on object relevance to support user context. Our findings show how the proposed design principles and metrics can implement and assess alignment with user preferences, providing a foundation for implementing choice functions in IUIs that accurately reflect user preferences.
\end{enumerate}
The remainder of this paper is organized as follows: Section~\ref{sec:background} provides the necessary background. In Section~\ref{sec:related}, we review related work, highlighting the limitations of existing methods. Section~\ref{sec:extract} presents our proposed design principles, and Section~\ref{sec:metrics} introduces the evaluation metrics, detailing how they enable the assessment of alignment with preferences. Section~\ref{sec:experiments} describes the empirical study conducted with an automotive assistant, including methodology and results. Section~\ref{sec:discussion} discusses the findings, limitations of our framework, and outlines directions for future research. Section~\ref{sec:impact} reflects on the framework's societal impact. Finally, Section~\ref{sec:conclusions} concludes the paper.

\section{Background}
\label{sec:background}
For IUIs to make context-aware, personalized, and intelligent offers or decisions, they must rely on a robust model of (user) preference. In this paper, we will study a very general representation of preferences, i.e., a binary relation over a set of alternatives. These alternatives could range from different layout options for an interface, various response strategies for a digital assistant, to commands to control a smart device or feature. Let $X$ denote the set of all possible alternatives available.

A \emph{preference (ordering)}~$\succsim$ is a binary relation over~$X$, where $x \succsim y$ denotes that alternative~$x$ \emph{is at least as preferred as} alternative~$y$. Preferences are \emph{rational} when they satisfy the following properties:
\begin{enumerate}
  \item \emph{Completeness}: For all $x, y \in X$, either $x \succsim y$ or $y \succsim x$, or both.
  \item \emph{Transitivity}: For all $x, y, z \in X$, if $x \succsim y$ and $y \succsim z$, then $x \succsim z$.
\end{enumerate}
Intuitively, rational preferences provide a notion of a ranking among alternatives, in which some alternatives may be tied. Mathematically, rational preferences are a weak order. For convenience, we assume user preferences to be rational, but explicitly indicate when this assumption is relaxed.

We are often interested in \emph{strict preferences}~$\succ$ and \emph{indifference}~$\sim$. These derived concepts can be defined from~$\succsim$. We say that $x \succ y$ iff $x \succsim y$ and not $y \succsim x$, i.e, one alternative is strictly more preferred than the other. We say that $x \sim y$ iff $x \succsim y$ and $y \succsim x$, i.e., an agent (or user) is indifferent between the alternatives. It follows from the definition that~$\succ$ is a strict weak order, and~$\sim$ is an equivalence relation. We will refer to the disjoint subsets of~$X$ formed by the equivalence classes of~$\sim$ as \emph{indifference sets}.
Only a very special case of rational preferences is a \emph{ranking}, where for all $x, y \in X$, either $x \succ y$ or $y \succ x$, but not both. Rankings have a unique most preferred alternative in any subset of alternatives. In general, however, rational preferences can be more complex and may have multiple most preferred alternatives in any subset of~$X$.

A construct that can operationalize preferences is the choice function. A \emph{choice function}~$c$ maps every non-empty subset of alternatives~$A \subseteq X$ to (the chosen) alternatives~$c(A) \subseteq A$. We say that~$c$ \emph{encodes} or \emph{represents} the preference~$\succsim_c$ defined by $x \succsim_c y$ iff $x \in c(\{x,y\} \cup A)$ for some~$A \subseteq X$. In other words, $x \succsim_c y$ whenever $c$ chooses $x$ in presence of $y$. We say that~$c$ is \emph{rational} whenever~$\succsim_c$ is rational.

Many IUIs implement choice functions. For example, a recommendation system may choose the most relevant items from a set of options, a chatbot may select the most appropriate response from a set of possibilities, or a smart home system may choose the most suitable actions from a set of available alternatives. In this paper, we focus on IUIs that implement choice functions using \emph{large language models}~(LLMs).
An LLM is a neural network model that is trained on a large corpus of text data, designed to process and generate natural language based on input prompts. Integrated into an IUI, with prompts encoding instructions such as device control, navigation assistance, or content recommendations, LLMs can generate outputs that represent offers, decisions, or actions. By further integrating context and user-specific data, the LLM can offer suggestions or perform decisions tailored to the user. In other words, LLMs can implement choice functions in IUIs.
However, state-of-the-art LLMs are currently sensitive to the natural language formulation of instructions and, in particular, the object order of objects~(cf.~\cite{hou2023}). Hence, a choice function implemented using LLMs may not be rational unless special considerations are taken.
In Section~\ref{sec:extract}, we put forward design patterns for using LLMs to implement rational choice functions.

In particular, we are interested in choice functions that are aligned with user preferences. In this paper, we distinguish two types of alignment: partial alignment and full alignment. To begin with, suppose we have a user preference~$\succsim_u$ and a choice function~$c$. We say that~$c$ is \emph{partially aligned} with~$\succsim_u$ if for all $x, y \in X$, $x \succ_u y \implies x \succ_c y$. In other words, a choice function is partially aligned with user preferences if it never violates the user's strict preferences. We say that~$c$ is \emph{fully aligned} with~$\succsim_u$ if for all $x, y \in X$, $x \succ_u y \iff x \succ_c y$. In other words, a choice function is fully aligned with user preferences if it never violates the user's strict preferences or indifference relations. Hence, in addition to partial alignment, full alignment requires respecting indifference relations. (In fact, it is easy to see that the axioms $x \succ_u y \impliedby x \succ_c y$ and $x \sim_u y \implies x \sim_c y$ are equivalent if $c$ is partially aligned with $\succsim_u$.)

We assume that it is generally challenging to implement choice functions that align with user preferences in practice. An otherwise well-aligned choice function might violate the relevant alignment axioms for some pair $(x, y)$ of alternatives~$x,y \in X$. Formally, the pair $(x, y)$ is \emph{discordant} if $x \succ_u y$ and $y \succ_c x$ (i.e., the user strictly prefers $x$ over $y$ but the choice function $c$ prefers $y$ over $x$), and \emph{weakly discordant} if $x \succ_u y$ and $x \sim_c y$ or $x \sim_u y$ and $x \succ_c y$ (i.e., the user strictly prefers $x$ over $y$ but the choice function $c$ is indifferent between $x$ and $y$, or vice versa). However, the pair $(x, y)$ is \emph{concordant} if $x \succ_u y$ and $x \succ_c y$. We put forward metrics that quantify partial and full alignment in Section~\ref{sec:metrics}.

\section{Related Work}
\label{sec:related}
Related work proposes the use of LLMs to implement choice functions that rank alternatives, in particular documents and text passages in the domain of information retrieval.

\paragraph{Pointwise Ranking}
Several approaches use LLMs to assign a score to each individual alternative (pointwise). The method in~\cite{liang2023} calculates scores based on the probability assigned by the LLM to the query given each document. Similarly, \cite{sachan-etal-2022} employs log probabilities of the output in a binary classification task. An alternative approach is to directly prompt the model to rate relevance~\cite{ji2023}. All three approaches then use the scores to produce a ranking among alternatives. An avenue to generalize from rankings to rational preferences could be to arbitrarily partition the range of scores into discrete bins. However, \cite{liang2023} and~\cite{sachan-etal-2022} require white-box access to log probabilities of the LLM, limiting their applicability with proprietary (black-box) models. In contrast, the approach by~\cite{ji2023} can be adjusted to produce rankings with ties by instructing an LLM to assign an integer score within bounded intervals. We will refer to this design principle as~\texttt{Pointwise}. Recall that a ranking with ties expresses rational preferences that allow for strict preferences and indifference among alternatives. Ties are guaranteed to occur whenever the cardinality of scores is lower than the number of alternatives. 

\paragraph{Listwise Ranking}
In contrast to pointwise ranking, which evaluates individual alternatives, listwise techniques~\cite{sun2023chatgpt,ma2023} operate on a list of all available alternatives; that is, the model re-ranks the alternatives from the input list in a single step, without the need for producing intermediate scores for individual elements.
In practice, however, LLMs are sensitive to the order of alternatives provided in their prompt~\cite{hou2023}, leading to different orderings when the order of alternatives is permuted in the input. Moreover, LLMs often fail to reproduce all alternatives, invent new alternatives not included in the input, or repeat items~\cite{sun2023chatgpt}. This makes listwise approaches less reliable. Most importantly, however, listwise approaches do not represent indifference between alternatives, as they produce ranking without ties. We will refer to this design principle as~\texttt{Listwise}.

\paragraph{Pairwise Ranking}
An avenue to address the sensitivity of LLMs to the order of alternatives is to repeatedly query the model with permuted input, and ironing out inconsistencies in a postprocessing step. For instance, \emph{pairwise ranking prompting}~(PRP, \cite{qin2023}) leverages comparisons of document pairs. When comparing two documents, the LLM is instructed twice, once for each order of presentation. If it consistently prefers one alternative over the other, a strict preference relationship between the alternatives is established; otherwise, they are considered equally preferred. In order to aggregate from pairwise comparisons to a ranking among all alternatives, two relevant ranking methods are proposed in~\cite{qin2023}: \emph{PRP-Allpair} and \emph{PRP-Sort}. \emph{PRP-Allpair} aggregates scores from pairwise comparisons, where a document receives one point for each alternative it is strictly preferred to, and half a point for each alternative with which it is equally preferred, with fallback rankings for ties. \emph{PRP-Sort} gradually improves an initial ranking by swapping the positions of alternatives, but is not guaranteed to reach a unique fix-point. Similar to pointwise and listwise methods, pairwise ranking methods do not account for indifference between alternatives. Nonetheless, in Section~\ref{sec:extract}, we adopt the concept of pairwise prompting and integrate the idea to score alternatives from \emph{PRP-Allpair} to express rational preferences that include indifference.

\paragraph{Ranking Evaluation}

When evaluating rankings against some ground truth in the domain of information retrieval, attention is often limited to the top-k documents in any ranking. In turn, binary classification metrics such as precision, recall and F1 scores are commonly used to assess retrieved and ranked documents. More sophisticated metrics also exist in information retrieval, e.g., based on the idea of \emph{discounted cumulative gain}~\cite{ndcg,katerenchuk2018}, which operate on ground truth relevance scores and discount these scores by their position in the ranking. 
In the more general domain of preferences, however, we neither have binary labels indicating significant preference, nor do we have scores expressing how preferred an alternative might be. We only have relative propositions.

Another relevant metric is \emph{gamma coefficient}~\cite{gamma_coeff} which quantifies the similarity between two rankings by calculating the ratio between the difference and the sum of concordant and discordant pairs. 
However, the gamma coefficient requires strict rankings and ignores ties or indifference between alternatives.

\vskip 1em
In conclusion, related work in the literature addresses the problem of ranking alternatives, particularly documents. We will take inspiration from PRP in the following section, and use slight adjustments to pointwise and listwise ranking methods as baselines in our empirical study reported in Section~\ref{sec:experiments}. Rational preferences are a more general concept than rankings and require different, less specialized metrics.

\section{Design Principles for Implementing Rational Choice Functions with LLMs}
\label{sec:extract}
In this section, we propose a set of design principles for using LLMs to implement choice functions that represent rational preferences. Following the idea from~\cite{qin2023} we apply pairwise prompting over a finite set of alternatives~$X$, querying an LLM about which of the alternatives $x$ or $y$ is strictly preferred in a given context for all pairs~$x,y \in X$ with~$x \neq y$. Formally, we characterize the corresponding output of the LLM as a mapping~$Q$ from the pair~$(x,y)$ to the set~$\{x, y, \diamond, \bot\}$, where $\diamond \not\in X$ is a symbol for the model indicating indifference, and $\bot \not\in X$ is a symbol indicating the model's failure to produce $x$, $y$ or~$\diamond$ (i.e., a response that is \emph{not valid}). We define the \emph{queried preference}~$\succsim_Q$ by~$x \succsim_Q y$ iff $Q(x,y) \in \{x, \diamond\}$ or $Q(y,x) = \{x, \diamond\}$. Accordingly, we derive the \emph{queried strict preference}~$\succ_Q$ and the \emph{queried indifference}~$\sim_Q$.
Note, however, that~$\succsim_Q$ is not guaranteed to be rational, given the sensitivity of LLMs to the ordering of objects in their prompts. In fact, $\succsim_Q$ might not be complete, for instance, when $Q(x,y) = \bot$ and $Q(y,x) = \bot$ for some~$x,y \in X$ with~$x \neq y$.
Even if we assume that~$Q$ never maps to~$\diamond$ or~$\bot$, any choice function that encodes~$\succsim_Q$ might not be rational, since the queried preference can violate transitivity.

\paragraph{Pairwise Scoring}
Assuming that~$Q$ never produces responses that are not valid~(i.e., $\bot$), the queried relation~$\succsim_Q$ becomes complete. Then, by utilizing the the concept of scoring from PRP~\cite{qin2023}, we can obtain a preference that is rational. First, we assign a score to each~$x \in X$ according to Equation \ref{tab:scoringfunc}.
\begin{equation}
\texttt{score}(x) = |\{x \succ_Q y \mid y \in X\}| + 0.5 * |\{ x \sim_Q y \mid y \in X\}|
\label{tab:scoringfunc}
\end{equation}
In words, we compute the score of $x \in X$ by counting the alternatives~$y \in X$ for which~$x$ is strictly more preferred than~$y$ and adding half a point for each element in the indifference set containing~$x$.
We can define a choice function~$c$ by
\begin{equation}
    c(A) = \underset{x \in A}{\mathrm{arg\,max}}\  \texttt{score}(x)
    \label{tab:choicefunc}
\end{equation}
It is easy to see that $\succsim_c$ is rational, because the score induces a ranking with ties. This is in contrast with tie-breaking in \emph{PRP-Allpair}~\cite{qin2023}, as it explicitly allows for indifference. We refer to this pairwise scoring approach for implementing rational choice functions using LLMs as \texttt{Pairwise-Score}.

\paragraph{Component-based Method}
Instead of relying on scoring alternatives to implement a rational choice function, we propose enforcing transitivity by careful analysis of the queried preference. Specifically, we construct a directed graph~$G$ from $\succsim_Q$ with vertices representing each~$x \in X$ and edges directed from~$x$ to~$y$ whenever~$x \succsim_Q y$. If $\succsim_Q$ is complete, the graph is connected. We can then implement a choice function by computing \emph{strongly connected components} (SCC) of~$G$, partitioning the set of alternatives into disjoint subsets. For~$x \in X$ we denote by SCC($x$) the set of alternatives within the same SCC as~$x$. Trivially, each indifference set is contained within an SCC, and two indifference sets belong to different SCCs whenever edges in~$G$ are directed exclusively from alternatives in one indifference set to the other. Hence, the SCCs in~$G$ induce a relation on indifference sets representing a rational preference. We exploit this observation to define a rational choice function~$c$ as follows.
\begin{equation}
c(A)=\{ x \in A \mid \forall y \in A, x \neq y: (x \succsim_Q y) \lor x \in \texttt{SCC}(y)\}
    \label{tab:choice_scc}
\end{equation}
We refer to our component-based approach as \texttt{Pairwise-SCC}.

\section{Measuring Alignment with Preferences}
\label{sec:metrics}
Having established a selection of design principles to aid the implementation of choice functions that can represent rational preferences, we now turn our attention to quantifying alignment with user preferences in terms of partial and full alignment.

\paragraph{Partial Alignment}
Given a set of alternatives~$X$, a choice function~$c$ and user preference~$\succsim_u$ defined on~$X$, we propose to quantify partial alignment by calculating the ratio of strict preferences that are not violated by~$c$. We call this ratio the \emph{strict preference overlap}~(SPO). With a slight abuse of notation, we associate with~$\succ_u$ the set of all pairs~$(x,y)$ such that~$x \succ_u y$ for~$x, y \in X$. Then, we define SPO as follows.
\begin{equation}
\text{SPO}(c,\succsim_u) = 1 - \frac{|\succ_u - \succ_c|}{|\succ_u|}
\label{tab:halfjaccard}
\end{equation}
Observe that higher values indicate better alignment. In fact, $\text{SPO}(c,\succsim_u) = 0$ if the choice function violates all strict user preferences; e.g., for every pair $x,y \in X$, if $x \succ_u y$ then $y \succsim_c x$. SPO reaches its maximum value~1 when the choice function satisfies all strict user preferences. thus fulfilling the partial alignment axiom. 

\paragraph{Full Alignment}
To quantify full alignment, i.e. alignment of both strict preferences and indifference, we propose the use of \textit{Kendall distance with penalty} \Kp~\cite{fagin} as shown in Equation~\ref{tab:kendall}. Given a set of alternatives~$X$, a choice function~$c$ and user preference~$\succsim_u$ defined on~$X$, this metric counts discordant pairs and, weighted by a constant~$p$ with~$0 \leq p \leq 1$, the weakly discordant pairs. Let~$D = \{ (x,y) \in X \mid (x \succ_u y) \land (y \succ_u x) \}$ and $W = \{ (x,y) \in X \mid (x \succ_u y) \land (x \sim_c y) \lor (x \sim_u y) \land (x \succ_c y) \}$, then
\begin{equation}
\text{\Kp}(\succsim_u, \succsim_c) =|D|+ p*|W|
\label{tab:kendall}
\end{equation}
Note that, unlike the commonly used \emph{Kendall tau distance}, which considers only discordant pairs~$D$, \Kp~also incorporates weakly discordant pairs~$W$. The parameter~$p$ should be chosen carefully, depending on the goals of the application. Since \Kp is a distance measure, lower values indicate better alignment, and \Kp equals zero when the choice function~$c$ is fully aligned with~$\succsim_u$.

\section{Empirical Study}
\label{sec:experiments}
To demonstrate the applicability of our framework, we report results of an empirical study related to a use case from the automotive industry. The use case involves an IUI such as a voice assistance system, utilizing object detection data to tailor its decision-making and provide more engaging and context-aware responses. The primary goal is to correctly select the most contextually relevant object from an arbitrary set, according to user preferences. For instance, if a driver wishes to navigate to a park while it is raining at the destination and the system does not detect an umbrella in the passenger vehicle, the assistant could remind the driver to pack one. If a rain coat is present, however, the assistant should recognize that the user is indifferent to packing an umbrella and could commend the driver for bringing a rain coat. In fact, the IUI must implement a choice function that reflects user preferences among all candidate items based on their relevance to the context.

In our analysis, we compare implementations of the design principles \texttt{Pairwise-Score} and \texttt{Pairwise-SCC} using the LLMs GPT-4~\footnote{gpt-4-1106-preview}~\cite{openai2023gpt4}, Gemini\footnote{Gemini Pro as released on December 6, 2023}~\cite{team2023gemini}, and Llama2~\footnote{4-bit quantization of Llama-2-70b-chat}~\cite{touvron2023llama}, in terms of alignment with the preferences of a model user. To allow comparison with adjustments made in related work, our analysis also includes implementations of \texttt{Pointwise} and \texttt{Listwise}.
We also consider a na\"ive implementation of a choice function that simply queries for strict preferences called \texttt{Pairwise-Test}. It is assumed that in absence of proper design principles, many choice functions are implemented with LLMs follow this approach. \texttt{Pairwise-Test} is not guaranteed to yield rational choices, but we can still include it into an analysis of partial alignment.

Albeit the LLMs included in our study represent the state-of-the-art at the time of our experiments, their performance can be sensitive to the natural-language formulation of instructions. We have experimented with prompt engineering, and will report results obtained with the best-scoring templates indentified in this paper. We also allow \textit{none} as a valid response in the case of \texttt{Pairwise-Score} and \texttt{Pairwise-SCC} as outlined in Section~\ref{sec:extract}, indicating indifference~($\diamond$).
Occasionally, an implementation still fails to produce a choice, for instance when the output of the underlying LLM does not adhere to the required schema~(cf. $\bot$ from Section~\ref{sec:extract}). Thus, we report validity metrics to ensure a fair comparison. For transparency, we have included relevant prompt templates implementing the considered principles in the appendix of this paper.

Our study considers 23 contexts, including navigation destinations and weather conditions, and 40 personal objects that may be found inside a passenger vehicle. For each context, the preferences of our model user comprise up to four indifference sets among objects. Where applicable, we evaluate the performance of all implementations regarding partial alignment with user preferences in terms of SPO and full alignment with user preferences in terms of \Kp with the penalty parameter set to~$p=0.5$ (i.e., penalizing weakly discordant pairs at $50\%$ of the rate for discordant pairs).

\paragraph{Partial Alignment}
We begin by assessing partial alignment results obtained by using SPO.
\begin{table*}
  \caption{Average partial alignment scores across contexts for which the LLM produced valid responses using listed template. Higher scores are better.}
  \label{tab:partial}
  \centering
  \begin{tabular}{lccccccccc}
    \toprule
        & \multicolumn{3}{c}{\textbf{GPT-4}} & \multicolumn{3}{c}{\textbf{Gemini}} & \multicolumn{3}{c}{\textbf{Llama2}}\\
    \cmidrule(lr){2-4}\cmidrule(lr){5-7}\cmidrule(lr){8-10}
        & score & valid & template & score & valid & template & score & valid & template \\
    \midrule
        \textbf{\texttt{Pairwise-Score}} & \textbf{0.85} & \textbf{23/23} & T\ref{template:4_1} & \textbf{0.84} & \textbf{23/23} & T\ref{template:5_1} & \textbf{0.68} & \textbf{23/23} & T\ref{template:2_2} \\
        
        \textbf{\texttt{Pairwise-SCC}} & 0.14 & \textbf{23/23} & T\ref{template:6_1} & 0.11 & \textbf{23/23} & T\ref{template:5_1} & 0.01 & \textbf{23/23} & T\ref{template:2_2} \\
        
        \textbf{\texttt{Pairwise-Test}} & 0.80 & 21/23 & T\ref{template:3_1} & 0.76 & \textbf{23/23} & T\ref{template:1_1} & 0.21 & \textbf{23/23} & T\ref{template:2_2} \\ 
    \midrule
        
        \textbf{\texttt{Pointwise}} & 0.67 & \textbf{23/23} & T\ref{template:p2_2} & 0.58 & \textbf{23/23} & T\ref{template:p3_4} & 0.54 & \textbf{23/23} & T\ref{template:p3_4} \\
        
        \textbf{\texttt{Listwise}} & 0.94 & 7/23 & T\ref{template:l2_2} & 0.83 & \textbf{2/23} & T\ref{template:l1_4} & - & 0/23 & - \\ 
    \bottomrule
  \end{tabular}
\end{table*}
A summary of our experimental results is shown in Table~\ref{tab:partial}. Starting with baselines from related work, our results show that \texttt{Listwise} rarely produces a valid response because the underlying LLM does not adhere to the required schema, particularly, recalling all objects in a ranking. For instance, template T\ref{template:l2_2} with GPT-4 produced rational choices in only~7 of~23 contexts amongst all available objects, while Llama2 failed on all contexts, indicating limitations of this approach. In contrast, \texttt{Pointwise} proved to be more effective in representing rational preferences, making it comparable to our proposed design principles. In fact, \texttt{Pairwise-Score} and \texttt{Pairwise-SCC} achieved comparable results in terms of valid responses. However, a discrepancy is observed in the performance of \texttt{Pairwise-Test}, which fails in~2/23 contexts when using GPT-4 but was successful when using Gemini and Llama2. When projecting our analysis onto LLMs, our results highlight the influence of model choice on performance, with Llama2 generally performing worse than GPT-4 and Gemini in our study. In terms of partial alignment, \texttt{Pairwise-Score} emerged as the best-performing method, narrowly surpassing even \texttt{Pairwise-Test} by a small margin. \texttt{Pairwise-SCC} underperformed, exhibiting low scores across all models. Upon closer inspection, we found that \texttt{Pairwise-SCC} yielded only~2.04 (GPT-4 and Gemini: 2.04; Llama2: 1.04) indifference sets, whereas \texttt{Pairwise-Score} reported~21.3 (GPT-4: 24.5; Gemini: 18.4), and \texttt{Pointwise} reported~3.91 (GPT-4: 4.00; Gemini: 3.92). For comparison, the model user’s preferences comprise an average of~3.56 indifference sets. In our application scenario, constructing indifference sets via SCCs from queried preferences (\texttt{Pairwise-SCC}) results in very few and overly large indifference sets -- leading to more violations of the user's strict preferences. This is undesirable. On the other hand, \texttt{Pairwise-Score} produces more indifference sets than any of the other methods -- including more than the model user. In other words, \texttt{Pairwise-Score} represents rational preferences that are stricter than the user's. However, this is consistent with the goal of partial alignment. In fact, SPO does not penalize choice functions that represent stricter preferences than the user's. Somewhat unexpectedly, \texttt{Pairwise-Score} performs better than the na\"ive \texttt{Pairwise-Test} across all models, the most pronounced difference observed for Llama2. This suggests that pairwise prompting, c combined with inconsistency resolution as per our design principle, yields a choice function more aligned with user preferences. Hence, according to our study, \texttt{Pairwise-Score} is the preferred principle when partial alignment suffices. Using GPT-4 and prompt template~T\ref{template:4_1}, \texttt{Pairwise-Score} violates only $15\%$ of the model user's strict preferences in an average context.

\paragraph{Full Alignment}
We now turn our attention to full alignment, as measured by~\Kp. Recall that \Kp is not a normalized measure, which makes it somewhat more difficult to interpret the results without a reference. Clearly, distance is zero when a choice function is fully aligned with user preferences. At the other extreme, if a choice function were to represent an inversion of user preferences, i.e., $y \succsim_c x$ whenever $x \succsim_U y$, the full alignment score in our application scenario averaged $568$ across the 23 contexts. Note, however, that technically inverting user preferences does not necessarily constitute the worst-case score (e.g., inverting indifference between two alternatives still results in indifference). Another upper bound is the number of alternative pairs, calculated as $40^2 = 1600$.

\begin{table}
  \caption{Average full alignment scores across contexts for which the LLM produced valid responses using listed template. Lower scores are better.}
  \label{tab:full}
  \centering
  \begin{tabular}{lccccccccc}
    \toprule
        & \multicolumn{3}{c}{\textbf{GPT-4}} & \multicolumn{3}{c}{\textbf{Gemini}} & \multicolumn{3}{c}{\textbf{Llama2}}\\
    \cmidrule(lr){2-4}\cmidrule(lr){5-7}\cmidrule(lr){8-10} \\
        & score & valid & template & score & valid & template & score & valid & template \\
    \midrule
    \textbf{\texttt{Pairwise-Score}} & 241.5 & \textbf{23/23} & T\ref{template:6_1} & 253.5 & \textbf{23/23} & T\ref{template:5_1} & 254.1 & \textbf{23/23} & T\ref{template:5_2} \\
    \textbf{\texttt{Pairwise-SCC}} & \textbf{170.7} & \textbf{23/23} & T\ref{template:4_1} & \textbf{167.4} & \textbf{23/23} & T\ref{template:5_1} & \textbf{176.9} & \textbf{23/23} & T\ref{template:2_2} \\
    \midrule
    \textbf{\texttt{Pointwise}} & 191.6 & \textbf{23/23} & T\ref{template:p1_4} & 208.0 & \textbf{23/23} & T\ref{template:p1_4} & 182.1 & \textbf{23/23} & T\ref{template:p4_2} \\
     \bottomrule
\end{tabular}
\end{table}
A summary of our experimental results on full alignment is shown in Table~\ref{tab:full}, limited to design principles that produced valid responses across all tested contexts.
Minor changes to the prompt templates -- compared to those used in the partial alignment experiments -- slightly improved scores. Statistics on indifference set remained largely unchanged, however. We found that \texttt{Pairwise-SCC} still produced only~1.65 indifference sets (GPT-4 and Gemini: 2.04; Llama2: 1.04), while \texttt{Pairwise-Score} reported~20.7 (GPT-4: 24.5; Gemini: 6.56), and \texttt{Pointwise} reported~3.95 (GPT-4: 3.95; Gemini: 3.34).
Turning our attention to the scores, \texttt{Pairwise-SCC} achieved the closest alignment to user preferences, exhibiting the lowest (distance) scores among all methods tested.
In our best setting, we observed an average \Kp of 167.4. Considering $n=40$ alternatives (i.e., preferences expressed over $O(n^2)$ pairs), \texttt{Pairwise-SCC} avoids violation of the full alignment axioms for the vast majority of pairs -- through some room for improvement remains.
These findings suggest that under conditions requiring full alignment, \texttt{Pairwise-SCC} is the preferred choice for preference representation. The method demonstrates a consistent performance gradient across LLMs, slightly outperforming \texttt{Pairwise-Score} and \texttt{Pointwise}.
It is noteworthy that, when projecting our analysis onto LLMs, Llama2 performs comparably to GPT-4 and Gemini across all methods. None of the models consistently outperform the others in achieving full alignment -- unlike the more varied results observed in the partial alignment setting.

\section{Discussion, Limitations and Future Work}
\label{sec:discussion}
The results of our empirical study demonstrate the potential of implementing a choice function for use in IUIs based on LLMs when guided by specific design principles. Merely querying an LLM for strict preferences (as evidenced by \texttt{Pairwise-Test}) does not robustly represent user preferences. The reasons are manifold. For one, recent LLMs can be very sensitive to how instructions are formulated as prompts. For another, they are equally sensitive to the order in which objects appear in the prompt. In this paper, we have proposed design principles to address these challenges and to guide the development of LLM-based choice functions for IUIs. Our methods are inspired by approaches used in LLM-based document ranking but generalise from ranking (i.e., strict preferences over all candidates) to rational preferences. We also showed how some related methods can be adjusted to represent rational preferences, including \texttt{Pointwise}, which performed reliably, albeit not always competitively, in our empirical study.
Pairwise prompting methods can exhibit greater susceptibility to invalid or contradictory outputs from LLMs due to the larger number of candidate objects and, consequently, queries. However, depending on the strategy employed to resolve inconsistencies -- as shown in our proposed design principles -- contradictory outputs can be remediated to represent rational preferences, leading to distinct behavioral patterns. For example, the \texttt{Pairwise-SCC} principle tended to yield preferences with fewer indifference sets, reflecting many instances of tied or equally preferred alternatives. In contrast, the \texttt{Pairwise-Score} approach produced a more fine-grained preference structure, characterized by a higher number of indifference sets, thereby capturing subtler distinctions in preferences. These trends can be leveraged to implement more robust choice functions that align better with user preferences, either partially or fully.
Indeed, our empirical study further demonstrated the complementary strengths of different design principles. While \texttt{Pairwise-SCC} performed best in terms of full alignment with user preferences, it was less effective at capturing the nuances of partial alignment. Conversely, \texttt{Pairwise-Score} exhibited stronger partial alignment but was less effective in full alignment. These findings suggest that principles excel in one alignment context may not necessarily generalize well to others, and that the specific alignment goals of an application should guide the choice of design principle. This highlights the importance of aligning design principles with application-specific requirements for implementing choice functions with LLMs in IUIs.

However, several limitations of this study warrant discussion. First, our findings are based on a limited set of prompt templates, many of which resulted in invalid responses. As such, our results are influenced by choices made in prompt engineering. Additionally, the study used a dataset that does not assign weights on contexts or objects. In practice, some contexts or objects may appear more frequently than others in a given application. Both SPO and \Kp can be adjusted to incorporate weights for objects and object pairs. Evaluating an application's success in terms of how frequently an IUI makes correct choices, however, differs from measuring alignment with user preferences -- other metrics would be more appropriate for such assessments. Furthermore, it should be noted that the robustness offered by certain design principles -- particularly \texttt{Pairwise-SCC} and \texttt{Pairwise-Score} -- comes at the cost of added computational complexity (e.g., $O(n^2)$ where $n$ is the number of alternatives). Future research should focus on developing design principles that scale more efficiently and on refining prompt templates to enhance alignment across a broader range of contexts.

\section{Societal Impact}
\label{sec:impact}
The integration of LLMs within IUIs represents an important step toward advancing user-centric automation. In this context, aligning LLM-based decision-making with user preferences is critical, as it enhances the reliability and trustworthiness of these systems. Our contribution centers on design principles and metrics to assess alignment, helping ensure that IUIs reflect rational user preferences. This can improve user experience in areas where decision automation has tangible impacts, such as personalized services, smart home applications, and - as demonstrated in this paper - automotive assistance.

However, our framework of design principles and metrics also comes with limitations and ethical considerations. First, any conclusions drawn from applying proposed design principles and metrics are specific to the concrete implementation (e.g., prompt template) and the choice of LLM. While empirical results may be indicative of a model's general capacity to align with user preferences, developers should be wary of misinterpreting these results as evidence of broader alignment than what has been empirically validated, potentially leading to overconfidence in the LLM's reliability.
In addition, our framework addresses only rational preferences and provides no insight into other dimensions of alignment -- such as bias or toxicity -- which are inherently irrational but critical for responsible artificial intelligence. While preference alignment can be achieved, it should not be misconstrued as a comprehensive measure of alignment encompassing all ethical concerns.

\section{Conclusions}
\label{sec:conclusions}
In this paper, we presented a framework for implementing choice functions with LLMs for use in IUIs and for evaluating their alignment with user preferences. Addressing a key gap in alignment research, our approach extends beyond traditional ranking functions by encompassing the richer structure of rational preferences, which includes both strict preferences and indifference among alternatives. By introducing design principles that generalize from ranking to rational preferences, we have provided methods capable of representing nuanced user preferences in IUIs.

Our empirical study on an automotive use case demonstrates the framework’s applicability and highlights the distinct strengths of different design principles depending on whether full or partial alignment is desired. While methods such as \texttt{Pairwise-SCC} and \texttt{Pairwise-Score} show promise for capturing rational preferences in real-world applications, we observed that alignment quality is significantly influenced by model choice and prompt engineering. Furthermore, we found that current LLMs exhibit variable performance across alignment contexts, suggesting that the preferred design principle for choice functions should be chosen based on the specific alignment goals of the application.

These findings open several avenues for future research. One direction involves developing more scalable design principles to better support contexts with larger numbers of alternatives, as well as further refining prompt templates to improve alignment across diverse contexts. 

By ensuring that LLM-based decision-making reflects user preferences, our framework lays a foundation for more reliable and trustworthy AI-driven IUIs across a range of application domains. However, it is essential to view preference alignment as one component of a broader commitment to responsible AI -- requiring ongoing attention to other alignment dimensions such as fairness, bias, and transparency.

\appendix

\section{Prompt Templates}
\renewcommand{\lstlistingname}{Template}
\lstdefinestyle{prompt_template}
{
    basicstyle=\ttfamily\footnotesize,
    breaklines=true,
    breakindent=0pt,
    breakatwhitespace=true
}
\lstdefinestyle{userprompt}
{
    backgroundcolor=\color{paleaqua},frame=single, frameround=tttt,
    basicstyle=\color{black}\tffamily
}
We have employed instruction-tuned LLMs that follow a chat protocol, starting with a system-prompt, followed by a user-prompt. The prompt templates referenced in our empirical study are disclosed in this section, where the double dash~\texttt{-{}-} separates system-prompt (above) from user-prompt (below). The terms in (single) curly brackets are to be instantiated with object names (comma-separated if list of objects), natural language description of contexts, or bounds on integer scores, where appropriate. Double curly brackets are used to protect single curly brackets from replacement.
\begin{lstlisting}[style=prompt_template, label=template:4_1, caption=T\ref{template:4_1}]
You are a helpful assistant whose task is to decide which of the two objects is more relevant to a particular context. If neither is preferred over the other in the context, reply with none. Format the output as a JSON. Here is an example of the output format: {{"answer": ""}}. Do not provide any explanations.
--
Context: {context}\nObject 1: {object1}\nObject 2: {object2}\n
\end{lstlisting}
\begin{lstlisting}[style=prompt_template, label=template:5_1, caption=T\ref{template:5_1}]
You are a helpful assistant whose task is to decide which of the two objects (object 1 or object 2) is more relevant to a particular context. If neither is preferred over the other in the context, reply with none. Format the output as a JSON. Here is an example of the output format: {{"answer": ""}}. Do not provide any explanations.
--
Context: {context}\nObject 1: {object1}\nObject 2: {object2}\n
\end{lstlisting}
\begin{lstlisting}[style=prompt_template, label=template:2_2, caption=T\ref{template:2_2}]
You are a helpful assistant whose task is to decide which of the two objects (object 1 or object 2) is more relevant to a particular context. Format the output as a JSON. Here is an example of the output format: {{"answer": ""}}. Do not provide any explanations.
--
Object 1: {object1}\nObject 2: {object2}\nContext: {context}\n
\end{lstlisting}
\begin{lstlisting}[style=prompt_template, label=template:6_1, caption=T\ref{template:6_1}]
You are a helpful assistant whose task is to decide which of the two objects is more relevant to a particular context. If neither is preferred over the other in the context, reply with none. Format the output in JSON using the key "answer" and the name of the object as value. Do not provide any explanations.
--
Context: {context}\nObject 1: {object1}\nObject 2: {object2}\n
\end{lstlisting}
\begin{lstlisting}[style=prompt_template, label=template:3_1, caption=T\ref{template:3_1}]
You are a helpful assistant whose task is to decide which of the two objects is more relevant to a particular context. Format the output in JSON using the key "answer" and the name of the object as value. Do not provide any explanations.
--
Context: {context}\nObject 1: {object1}\nObject 2: {object2}\n
\end{lstlisting}
\begin{lstlisting}[style=prompt_template, label=template:1_1, caption=T\ref{template:1_1}]
You are a helpful assistant whose task is to decide which of the two objects is more relevant to a particular context. Format the output as a JSON. Here is an example of the output format: {{"answer": ""}}. Do not provide any explanations.
--
Context: {context}\nObject 1: {object1}\nObject 2: {object2}\n
\end{lstlisting}
\begin{lstlisting}[style=prompt_template, label=template:p2_2, caption=T\ref{template:p2_2}]
You are a helpful assistant whose task is to assign a score from {score_low} to {score_high} to an object based on its relevance to a particular context. Format the output in JSON using the key "score". Do not provide any explanations.
--
Context: {context}\nObject: {object}
\end{lstlisting}
\begin{lstlisting}[style=prompt_template, label=template:p3_4, caption=T\ref{template:p3_4}]
You are a helpful assistant whose task is to rate the relevance of an object in a particular context on a scale of {score_low} to {score_high}. Format the output in JSON using the key "score" and the assigned score as value. Do not provide any explanations.
--
Context: {context}\nObject: {object}\n
\end{lstlisting}
\begin{lstlisting}[style=prompt_template, label=template:l2_2, caption=T\ref{template:l2_2}]
You are a helpful assistant whose task is to rank objects based on their relevance to a particular context. Do not provide explanations. Provide the ranked objects as a list. Format the output in JSON using the key "ranking" and the list of ranked objects as value.
--
Context: {context}\nObjects: {objects}
\end{lstlisting}
\begin{lstlisting}[style=prompt_template, label=template:l1_4, caption=T\ref{template:l1_4}]
You are a helpful assistant whose task is to rank objects based on their relevance to a particular context. Do not provide explanations. Format the output in JSON using the key "ranking" and the list of ranked objects as value.
--
Context: {context}\nObjects: {objects}\n
\end{lstlisting}
\begin{lstlisting}[style=prompt_template, label=template:5_2, caption=T\ref{template:5_2}]
You are a helpful assistant whose task is to decide which of the two objects (object 1 or object 2) is more relevant to a particular context. If neither is preferred over the other in the context, reply with none. Format the output as a JSON. Here is an example of the output format: {{"answer": ""}}. Do not provide any explanations.
--
Object 1: {object1}\nObject 2: {object2}\nContext: {context}\n
\end{lstlisting}
\begin{lstlisting}[style=prompt_template, label=template:p1_4, caption=T\ref{template:p1_4}]
You are a helpful assistant whose task is to assign a score to an object based on its relevance to a particular context. Format the output in JSON using the key "score" and the assigned score from {score_low} to {score_high} as value. Do not provide any explanations.
--
Context: {context}\nObject: {object}\n
\end{lstlisting}
\begin{lstlisting}[style=prompt_template, label=template:p4_2, caption=T\ref{template:p4_2}]
You are a helpful assistant whose task is to assign a score from {score_low} to {score_high} to an object based on its relevance to a particular context. Format the output in JSON using the key "score". Here is an example of the output format: {{"score": "}}. Do not provide any explanations.
--
Context: {context}\nObject: {object}
\end{lstlisting}

\end{document}